# NIDA-CLIFGAN: Natural Infrastructure Damage Assessment through Efficient Classification Combining Contrastive Learning, Information Fusion and Generative Adversarial Networks

Jie Wei[1*], Zhigang Zhu[1], Erik Blasch[2], Bilal Abdulrahman[1], Billy Davila[1],

Shuoxin Liu[1], Jed Magracia[1], Ling Fang[1]

[1] Dept. of Computer Science, City College of New York   [2] Air Force Office of Scientific Research

∗ Point of Contact: E-mail: jwei@ccny.cuny.edu

## Abstract

During natural disasters, aircraft and satellites are used to survey the impacted regions. Usually human experts are needed to manually label the degrees of the building damage so that proper humanitarian assistance and disaster response (HADR) can be achieved, which is labor-intensive and time-consuming. Expecting human labeling of major disasters over a wide area gravely slows down the HADR efforts. It is thus of crucial interest to take advantage of the cutting-edge Artificial Intelligence and Machine Learning techniques to speed up the natural infrastructure damage assessment process to achieve effective HADR. Accordingly, the paper demonstrates a systematic effort to achieve efficient building damage classification. First, two novel generative adversarial nets (GANs) are designed to augment data used to train the deep-learning-based classifier. Second, a contrastive learning based method using novel data structures is developed to achieve great performance. Third, by using information fusion, the classifier is effectively trained with very few training data samples for transfer learning. All the classifiers are small enough to be loaded in a smart phone or simple laptop for first responders. Based on the available overhead imagery dataset, results demonstrate data and computational efficiency with 10% of the collected data combined with a GAN reducing the time of computation from roughly half a day to about 1 hour with roughly similar classification performances.

## 1 Introduction

### 1.1 Application Context

During natural disasters in populated regions, aircraft and satellites are used to survey the impacted regions, generating overhead imagery. From previous surveillance images of the same area, a comparison results in natural infrastructure damage assessment (NIDA). Usually human experts are needed to carefully inspect the damage levels of man-made infrastructures from these images and manually label the degrees of the damage. The qualitative NIDA includes categories of no damage, minor damage, major damage, and total damage. Using these categorical labels of NIDA, proper humanitarian assistance and disaster response (HADR) efforts can be properly planned and deployed

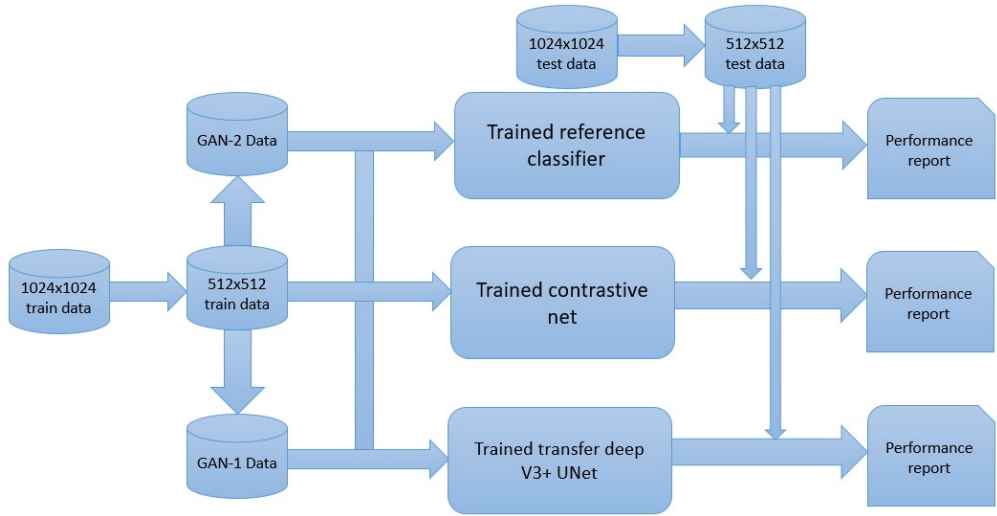

Fig. 1. Flowchart of the CLIFGAN approach

[4]. The manual labeling process is extremely labor intensive and time consuming as each individual building must be patiently studied to decide the correct labeling category wherein lots of expertise on the buildings and damage styles are needed; this trouble is even more dire for major disasters over a wide area when this process will gravely slow down the HADR efforts. Additionally, an exceedingly pressing concern within the HADR domain concerns the use of cutting-edge Artificial Intelligence and Machine Learning (AI/ML) techniques to automate the labeling process [10, 11, 12], which will significantly speed up the HADR delivery in a timely manner to better aid people in crisis; but this has been limited in interpretability and explainability for deployment, use, and sustainment. To ascertain the capability of AI/ML towards NIDA, the Efficient and Robust Machine Learning Center of Excellence (ERML COE) hosted an overhead imagery hackathon [8] as organized by the Air Force Research Lab, the Air Force Office of Scientific Research, the University of Wisconsin at Madison, and the Toyota Institute of Technology at Chicago. The overhead imagery hackathon leveraged the well-annotated overhead imagery dataset, xView. One of the main themes of this hackathon included data and computational efficiency advancements in NIDA. More specifically, building damage assessments were a main theme towards intelligent and agile image classifiers. With such data and computational efficiency, the methods could be used in smart phones and laptops to help HADR efforts. Our team participated the overhead imagery hackathon making a systematic effort to achieve efficient building damage classification using generative adversarial networks (GAN), contrastive learning, transfer learning and information fusion methods. Special care was taken to ensure the data and computation should be kept to minimal so that the overarching efficiency demand in HADR system can be guaranteed. The data and computational efficiency achieved is encouraging resulting in one of the three winners in the overhead imagery hackathon.

1.2  Motivations and Contributions

To achieve data and computation efficient damage assessment classification, the following 3 tasks were explored: 1) Develop two pix2pix GANs with novel architectures to augment data with desired size and nature; 2) Design contrastive learning for label-efficient semantic segmentation by training the model on limited data first to create a robust feature extractor using bespoke contrastive loss; and 3) Demonstrate Transfer learning, Unet and 2-level fusion using the newest DeepLab v3+ network initialized by mobilenet [9] to yield competitive performances with a small set of data.

The overhead imagery hackathon challenge focused on data and computational efficiency and robust performance without emphasis on the classification performance. Hence, the goal was to use the imagery dataset with reduced dimensions to improve the data and computation efficiency, while maintaining previous xView classification performances to ensure competitive approaches. In the previous challenge administered by Defense Innovation Unit (DIU) two years ago, there are several winning methods yielding great classification performances. By reproducing some of these entries as reference points, a comparative advantage was required over the baseline methods. Although only 2 years passed, the changes and evolution by python packages, such as pytorch and cuda, made our reproduction a great challenge. The lack of documentation and maintenance from previous entries resulted in difficulties to reproduce previous work. Eventually, two winning methods among the top 6 methods were fully reproduced. In Table 1, the performances of these 2 methods are reported as a reference of comparisons to our development.

The overall flowchart of the Contrastive Learning Information Fusion GAN (CLIFGAN) is shown in Fig. 1. The original image resolution of the xView dataset is 1024x1024, which is too large and computing demanding for our development. To render the CLIFGAN development viable, the images were reduced to 512x512 so that this dataset can be effectively analyzed by a laptop computer. As noted, two GANs are developed for low resource data augmentation. The GAN data and models are sent to the DL, CL, and TL classifiers for analysis; with subsequent classifier fusion. Since the data being used was imagery, then the results support CLIFGAN.

## 2  Two Novel Generative Adversarial Networks (GAN)

For current state-of-the-art models to localize the level-damage at buildings in the satellite images, pre-disaster and post-disaster pictures are crucial. However, a common problem encountered is the lack of labeled data. GANs can provide a solution to problems of unlabeled data for training, limited examples, and plausible scenarios [3]. The Pix2Pix GAN we used is a general approach for image-to-image translation [6]. The generator model of the GAN is provided with a given image as input and generates a translated version of the image. Meanwhile, the discriminator model is given an input image and a real or generated paired image and must determine whether the paired image is real or fake, resulting in a generator model trained to both fool the discriminator model and to minimize the loss between the generated image and the expected target image. The general architecture allows the Pix2Pix model to be trained for a range of image-to-image translation tasks such as generating a post-disaster image from a pre-disaster image. Here, two novel Pix2Pix GANs were developed to augment data with desired size and nature: GAN-1 and GAN-2.

### 2.1  GAN-1: Mask Controllable

GAN-1 (Mask Controllable GAN) is a Pix2Pix GAN that takes a 4-channel image and outputs a post-disaster image. GAN-1's input has four channels: the RGB channels of the pre-disaster image and one channel of the post-disaster labels. The output of the GAN would be the RGB channels of a generated post-disaster image. GAN-1 can generate as many post-disaster images as needed with desired labels. The workflow for GAN-1 works as follows. GAN-1 was trained on the xView dataset images that were reduced into 512x512 images. In order to speed up training time, the images were resized into 256x256 images during training and normalized the images. The GAN can predict different scenarios of destruction, whether it be floods, volcanoes, fires, hurricanes, etc. To create the new augmented data, existing pre-disaster images found in the xView dataset and the corresponding labels of post-disaster images were used as inputs. The labels were modified/controlled to get the desired destruction level and combined with the pre-disaster images as input for the GANs. The resulting combined images were passed into the GAN as inputs to generate the predicted post-disaster images.

### 2.2  GAN-2: Random Disaster Creator

GAN-2 (Random Disaster Creator GAN) is a Pix2Pix GAN that uses pre-disaster images and randomized damage level masks as input to generate output post-disaster images and masks. GAN-2's generator takes as input a 4-channel image which is composed of the pre-disaster image and an arbitrary post-disaster mask with a random damage level and the generator's output as a 4-channel image which can be split to generate the post-disaster image and its post-disaster mask. The workflow of GAN-2 is similar to GAN-1 but with certain differences. The pre-processing stage is the same as GAN-1. After the pre-processing stage, the Pix2Pix GAN-2 was trained by given as input a 4-channel image made up of the pre-disaster image and the real post-disaster mask and produce an output 4-channel image made up by the post-disaster image and its corresponding post-disaster mask. Once the Pix2Pix GAN generator was trained, just as for GAN-1, a function was designed to modify the damage level of the pre-disaster masks. A damage level was then randomly selected to produce an arbitrary post-disaster mask label and combine it with its pre-disaster image to generate the input 4-channel image. Once the input 4-channel image was obtained, it was sent to the trained Pix2Pix GAN-2 generator to create an output 4-channel image that can be split into two images resulting in the augmented data made of a post-disaster image of randomized damage level and the generated mask.

Table 1: Performances of 2 reproduced methods and CLIFGAN methods

|  | reproduced 1 | reproduced 2 | contrastive | transfer learning and fusion |
| --- | --- | --- | --- | --- |
| size | 228 MB | 441 MB | 40 MB | 9.7 MB |
| Segmentation F1 | 0.814 | 0.815 | 0.910 | 0.893 |
| Classification F1 | 0.614 | 0.497 | 0.674 | .664 |

## 3 Contrastive Learning

Contrastive Learning (CL) is a recent method to compare date for similarity and differences without labels. The CLIFGAN system employs contrastive pre-training for semantic segmentation from [13] and re-purposes it for segmentation of buildings. The CL technique pre-trains the feature extractor of the classifier model to predict similar features for objects in the images belonging to the same class. The CL model is then fine tuned for the final class prediction. The notable changes from the original method [13] include a mobilenet backbone pre-trained on the imageNet dataset for all experiments and compared with resnet-50. The use of mobilenet drastically reduces the size of the model as well as fewer parameters to train which results in faster training and less inference time. The trade-off of slightly lesser accuracy is worth the improvements in efficiency aligning more closely to the goals of the Overhead Imagery hackathon competition. The within image loss is used for the contrastive pre-training step.

To begin the CL process, the vanilla Deepnetv3+ model architecture [2] is trained with a pre-trained mobilenet backbone to use as a benchmark for the contrastive pre-training. Other approaches were experimented with such as dice loss, dice plus focal loss and cross-entropy loss. The best performance was achieved with the cross entropy loss. The data augmentations used across all the experiments were a combination of random scaling between scales 0.8 to 2, random cropping, and random flipping (horizontal and vertical). The initial learning rate used was 0.01 with a batch size of 16. The momentum used is 0.9; the weight decay is 1e-4; and a polynomial scheduler with power set to 0.9. The train dataset is divided into 2 parts: 90 % for training and 10 % for validation. The training was run until the evaluation performance on the validation set sees a continuous downward trend, i.e., the model starts over fitting the training data. The final model is trained for 344 epochs. For contrastive pre-training, the within-image loss without any distortions was selected as in the experiments, a reduction in performance when employing the color distortions [2] compared with no distortions. The final implementation proceeds by first adding a projection head to the deeplab v3+ model architecture similar to the original method, with three 1x1 convolution layers followed by a unit normalization layer.

After pre-training, The CL process discards the projection head and fine tunes the model using a cross entropy loss. The discard step is similar to the vanilla training. Everything else is set identical to the vanilla training. The system stops training when the validation sees a downward trend. Finally, by using the trained segmentation model in a siamese network architecture, classification of the post disaster buildings was achieved. As is the case in siamese networks, parameters are shared between the two branches. The predicted features from both images are first concatenated and then reduced to a single dimension using convolutional layers for destruction class prediction. As can be seen in Table 1, contrastive learning indeed achieves the best segmentation and classification performances over all methods we managed to implement.

Table 2: Performances of transfer learning and fusion for different data

|                  | Training time   | classification F1 | segmentation F1 |
|------------------|-----------------|-------------------|-----------------|
| Full data        | 11 hrs 23 mins  | 0.664             | 0.893           |
| 10% data         | 1 hr 2 mins     | 0.565             | 0.862           |
| 10% data + GAN-1 | 2 hrs 6 mins    | 0.592             | 0.874           |
| 10% data + GAN-2 | 2 hrs 8 mins    | 0.511             | 0.803           |

## 4 Transfer Learning and Information Fusion Methods

One of the challenges in deep learning deployment is to determine the best method. Instead of choosing the best method, robust approaches seek a fusion of classifiers. For the CLIFGAN method, the Unet is combined with transfer learning which has been exceedingly powerful in image segmentation [7]. Besides contrastive learning, it is thus another method to achieve data and computation efficient classification. Figure 3 shows example outputs of assessment. Using python's segmentation models and Matlab©'s segnetLayers methods on the xView datasets, all ended up with significantly worse results than the reference methods. The newest DeepLab v3+ network, recently made available by Matlab's deep learning toolbox, yield reasonably competitive results. The DeepLab v3+ net can be initialized by mobilenet, the smallest deep net with impressive performances, thus the valuable knowledge learned from ImageNet can be readily transferred to the disaster overhead images in the xView dataset. In this novel combination of Unet, DeepLab v3+, and Mobilenet architecture, depth separable convolutions are used in the atrous spatial pyramid pooling and decoding subnets, especially tailored for mobilenet. To further augment training, various scales and geometric transforms such as rotations, flipping, shearing are employed on the fly during training.

For user-machine fusion [1] motivation utilized 2-level image fusion improves the performances by: 1) independently train 3 different DeepLabv3+ nets and use the majority vote as the final label [2]; and 2) pixel level fusion: which uses mathematical morphological filtering [5] based procedure for post-processing to further enhance the results. The result delivered by network image fusion methods is also tabulated in Table 1. Its performance (0.664) is better than the 2 reproduced winning methods, and worse than the contrastive learning (0.674). However, the size of this method (9.7MB) is the smallest, among all four methods, which can be easily installed on smart phones and laptops for ease of use.

One unique benefit of this method is its robustness to the size and nature of the training data, while all other methods have trouble dealing with small data size and GAN-1 and GAN-2 augmented data: far worse results are generated for reasons we still need to thoroughly investigate; but this method can yield reasonably decent results. As shown in Table 2, with merely 10% of the xView training data, i.e., 280 training overhead images, the CLIFGAN net can still deliver a fairly good segmentation F1: compared with using the full data, the F1 only dropped 3%, from 0.893 to 0.862, with the training time reduced from half day to merely 1 hour. When using GAN-1 data, the classification F1 can be improved by 3%. This proved that GAN-1 can really help up better augment training data. Remember that a user of CLIFGAN can freely control the damage labels in GAN-1 to alleviate the annoying class unbalance problem. The data generated by GAN-2 doesn't seem to come from the same statistical performance, thus reducing both the classification and segmentation performances. It is interesting to observe that although GAN-2 generates visually appealing images, its blind use in classification could be a concern, as evidenced in our results tabulated in Table 2.

## 5 Conclusion

This paper describes the CLIFGAN approach for data and computation efficiency in NIDA as well as operational robustness. Specifically, data efficiency and computation efficiency were achieved by 1) significantly reducing the network size, by use of mobilenet in our system; 2) considerably cutting the need of training data, by using 512x512 images instead of 1024x1024 ones and furthermore a mere 10% training data can still yield acceptable classification performances; The CLIFGAN is operationally robust by the use of contrastive learning and image fusion in transfer learning, which can work with a relatively small training dataset; the use of novel Pix2Pix GANs, especially GAN-1, can further combat the lack of balanced training data. As a data and computation efficient and operationally robust approach, our CLIFGAN approach can be of great interest to NIDA and HADR community. Further work with the CLIFGAN on the xView dataset or other disaster and weather dataset includes conditioning on the types of disasters, multi-modal sensing and prediction [14], and focus on specific areas of interest and different types of NIDA, e.g., power plant and infrastructure destruction, which may not only improve the classification performance, but also could be of crucial interest to agile and efficient damage evaluation, human assistance and disaster relief efforts, and further global warming applications.

# 6 Appendices

Auxiliary figures helping better understand our work.

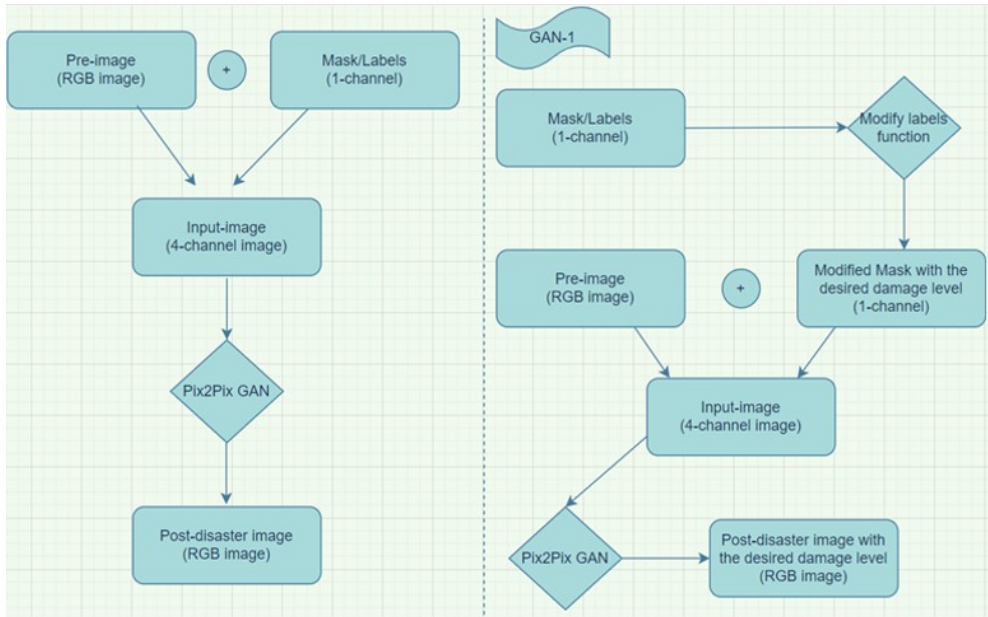

Appendix Fig. 1. Flowchart of GAN-1 & GAN-2

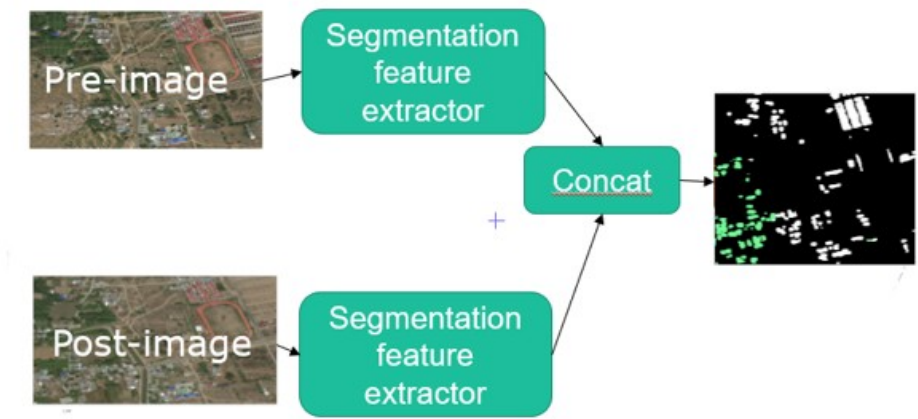

Appendix Fig. 2. Siamese network used for classification

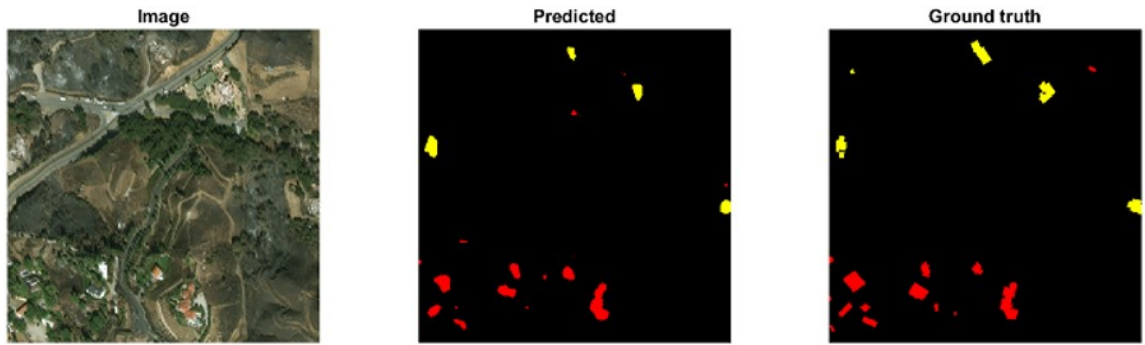

Appendix Fig. 3. Sample outputs by transfer learning Unet; red: no damage building, yellow: total destroyed buildings.